\documentclass{article} %

\PassOptionsToPackage{dvipsnames}{xcolor}

\usepackage[T1]{fontenc}
\usepackage{conference,times}
\usepackage{amsmath,amsfonts,bm}

\def\eqref#1{equation~\ref{#1}}

\def\1{\bm{1}}

\DeclareMathAlphabet{\mathsfit}{\encodingdefault}{\sfdefault}{m}{sl}
\SetMathAlphabet{\mathsfit}{bold}{\encodingdefault}{\sfdefault}{bx}{n}

\usepackage{enumitem}
\usepackage{float}
\usepackage{needspace}       %
\usepackage{tabularx}          %
\usepackage{multirow}
\usepackage{makecell}          %
\usepackage{tikz}
\usetikzlibrary{arrows.meta,backgrounds,calc,decorations.pathreplacing,fit,positioning}
\usepackage{xcolor}
\usepackage{amsmath}
\usepackage{subcaption}
\usepackage{graphicx}
\usepackage{wrapfig}
\usepackage{booktabs}
\usepackage[dvipsnames]{xcolor}

\definecolor{lightciteblue}{RGB}{146, 197, 222}
\definecolor{darkerblue}{HTML}{2244AA}
\definecolor{lightcitered}{HTML}{CC3366}
\usepackage[colorlinks=true, citecolor=darkerblue, urlcolor=lightcitered]{hyperref}

\newcommand{\revisioncolor}{\color{black}}
\newcommand{\revised}[1]{{\revisioncolor #1}}
\renewcommand{\revisioncolor}{}
\definecolor{toreviewblue}{RGB}{70, 150, 215}
\DeclareRobustCommand{\toreview}[1]{{#1}}
\definecolor{issuelight}{RGB}{230, 110, 110}
\definecolor{issuedark}{RGB}{160, 0, 0}

\newcommand{\panelref}[2]{\hyperref[#1]{\ref*{#1}#2}}

\definecolor{tocheckgreen}{RGB}{0, 140, 90}

\newif\ifshowinternal
\showinternalfalse
\definecolor{internalgray}{RGB}{170, 170, 170}
\DeclareRobustCommand{\internal}[1]{\ifshowinternal
  {\def\toreview##1{##1}\textcolor{internalgray}{#1}}\else\unskip\fi}
\usepackage{xspace}

\newcommand{\passone}{\text{Pass@1}\xspace}

\newcommand{\passfive}{\text{Pass@5}\xspace}
\newcommand{\seqone}{\text{Seq@1}\xspace}
\newcommand{\seqtwo}{\text{Seq@2}\xspace}
\newcommand{\seqthree}{\text{Seq@3}\xspace}
\newcommand{\seqfour}{\text{Seq@4}\xspace}
\newcommand{\seqfive}{\text{Seq@5}\xspace}
\newcommand{\seqk}{\text{Seq@K}\xspace}
\newcommand{\passk}{\text{Pass@K}\xspace}

\usepackage[most]{tcolorbox}

\definecolor{quotebg}{RGB}{240, 244, 248}      %
\definecolor{quoteborder}{RGB}{52, 95, 140}    %

\newtcolorbox{trajectory}{
  enhanced,
  breakable,
  width=\textwidth,
  colback=quotebg,
  colframe=quotebg,
  boxrule=0pt,
  arc=1pt,
  outer arc=1pt,
  left=10pt,
  right=8pt,
  top=6pt,
  bottom=6pt,
  borderline west={2.5pt}{0pt}{quoteborder},
  fontupper=\small
}

\definecolor{revlabel}{RGB}{98, 112, 122}
\newcommand{\revrunid}{}
\def\revfirsteight#1#2#3#4#5#6#7#8#9\relax{#1#2#3#4#5#6#7#8}
\newcommand{\runlink}[2]{}
\tcbset{
  revrun/.store in=\revrunid,
  revcaption/.style={before upper app={\xdef\revnumber{\thetcbcounter}},
    after={\par\vspace{-2pt}\noindent{\small\textbf{Example~\revnumber.}~#1}\par\vspace{8pt}}}
}
\newtcolorbox[auto counter, number within=section]{revexample}[3][]{
  revrun={},
  enhanced,
  breakable,
  width=\textwidth,
  colback=white,
  colframe=quoteborder!35,
  colbacktitle=quotebg,
  coltitle=quoteborder,
  boxrule=0.5pt,
  arc=3pt,
  outer arc=3pt,
  left=7pt,
  right=7pt,
  top=4pt,
  bottom=4pt,
  toptitle=3pt,
  bottomtitle=3pt,
  before skip=6pt,
  after skip=6pt,
  fonttitle=\sffamily\small\bfseries,
  fontupper=\small,
  before upper={\revfirsttrue},
  title={#2\enspace\textperiodcentered\enspace\mdseries #3%
    \ifx\revrunid\empty\else\internal{\enspace{\footnotesize run \revrunid}}\fi},
  #1
}
\newif\ifrevfirst
\newlength{\revlabelwidth}
\newcommand{\revfield}[2]{%
  \ifrevfirst\global\revfirstfalse\else
    \par\vspace{2pt}{\color{quoteborder!25}\hrule height 0.4pt}\vspace{4pt}\fi
  \par\noindent\begin{tabular}{@{}>{\raggedright\arraybackslash}m{\revlabelwidth}@{}m{\dimexpr\linewidth-\revlabelwidth\relax}@{}}
  {\sffamily\scriptsize\bfseries\color{revlabel}\MakeUppercase{#1}} & #2
  \end{tabular}\par\vspace{1pt}}

\usepackage{url}

\title{Seq@K: Evaluating Language Models Under Sequential Feedback}

\newcommand{\seqkauthors}{\parbox[t]{\dimexpr\textwidth-2\tabcolsep\relax}{\raggedright
Thomson~Yen\textsuperscript{1,*},
Gilbert~Yang\textsuperscript{1,*},
Naimeng~Ye\textsuperscript{1,*},
Minghui~Chen\textsuperscript{1,*},
Dong~Xia\textsuperscript{1},
Tianyi~Shen\textsuperscript{2},
Yechan~Jeon\textsuperscript{1},
Yaqin~Chen\textsuperscript{3},
Daksh~Mittal\textsuperscript{1},
Fangyu~Wu\textsuperscript{1},
Hongseok~Namkoong\textsuperscript{1}
\par\vspace{8pt}
{\normalfont\textsuperscript{1}Decision, Risk, and Operations Division, Columbia Business School,
\mbox{\textsuperscript{2}University of California, Santa Barbara},
\textsuperscript{3}Northwestern University\par\vspace{8pt}
\textsuperscript{*}Equal contribution}}}
\author{\seqkauthors}

\iclrfinalcopy %
\begin{document}

\maketitle

\begin{abstract}
    Language-model agents differ fundamentally from prior intelligent systems in their ability to learn from natural language in context.
This capability enables agents to iteratively refine their answers in response to feedback, a prevalent mode of operation in modern agentic workflows.
However, the community primarily ranks models by metrics such as \passone or \passk, which cannot capture how they utilize feedback on prior attempts.
We introduce \seqk, a natural evaluation protocol that measures whether a model can solve a problem within $K$ sequential attempts, provided feedback after each previous failure.
Unlike multi-turn agentic benchmarks that examine long-horizon planning, \seqk isolates and formalizes the core capability of refinement from feedback.
We evaluate Seq@K on a suite of {7 benchmarks and 16 frontier models}, and show that \seqk reveals a materially different model-selection landscape: under \seqk, (1) model rankings often differ,
(2) the Pareto frontiers admit more efficient configurations and (3) the gains from choosing the optimal model configuration can widen or shrink relative to those of \passk.
We further show that analyzing model trajectories under \seqk helps characterize iterative refinement behavior, such as the rate at which models address issues raised in feedback, or how rigidly they hold onto their own answers.
Finally, we study how the nature of feedback affects models' iterative refinement, showing that practitioners must carefully design and benchmark the feedback loops of iterative agentic systems.
Together, these results establish how \seqk offers a more faithful metric for intelligent systems that typically operate in such feedback loop, and the necessity to consider \seqk in evaluating models for these systems.

\end{abstract}
\section{Introduction}

Multi-turn interaction with feedback is a fundamental pattern of machine learning, driving classical paradigms including multi-armed bandits~\citep{auer2002finite}, active learning~\citep{settles2009active}, and reinforcement learning~\citep{sutton2018reinforcement}.
Recently, this same principle has catalyzed much of the progress in large language models (LLMs), underpinning advances in test-time compute~\citep{snell2024scaling}, automatic prompt optimization~\citep{yuksekgonul2025textgrad}, and agentic systems~\citep{park2023generative, yao2023react}.
Yet a striking disconnect has emerged between how these systems operate and how they are evaluated.
Even as modern methods~\citep{shinn2023reflexion, madaan2023selfrefine} and environments~\citep{wang2024mint, yao2024taubench} center on iterative refinement, the community still ranks models primarily by \passk~\citep{chen2021evaluating}, a strictly feedback-blind metric.

We argue that relying solely on feedback-blind evaluation is insufficient for developing agentic systems---it may reward the wrong behaviors and obscure the right ones.
For example, consider a coding agent: the user issues an initial request, the agent proposes an edit, the user responds with targeted critiques, and the agent revises.
This loop of proposal, feedback, and revision is the dominant pattern of agent usage today, with applications extending far beyond coding.
Yet \passk is blind to whether an agent can excel in this sequential decision-making loop.
This blind spot hides failure modes and undervalues models that successfully improve through iteration.

In this work, we quantify a model's ability to improve under sequential feedback by measuring it along a new axis, which we name \seqk.
\seqk measures whether a model solves a task within $K$ attempts when it receives feedback after each failed attempt.
Ranking models by \passk implicitly assumes that a model that is accurate in one turn also uses feedback well.
We show that this assumption often breaks down.
We evaluate \seqk on coding (MBPP Pro, HumanEval Pro~\citep{yu2025humanevalpro}), complex instruction following (AdvancedIF~\citep{he2025advancedif}), domain reasoning (CLBench DKR~\citep{dou2026cl}), and medical benchmarks (HealthBench~\citep{arora2025healthbench}).
Our results reveal how model rankings, cost frontiers, and failure modes change when models are allowed to revise their work based on feedback.

Our contributions and key empirical insights are structured around three primary questions:

\begin{enumerate}
    \item \revised{\textbf{Model Selection \& Cost Dynamics:} {Model rankings can change between evaluations based on \passk and \seqk. 
    We demonstrate (a) the benefit of \seqk by showing that it leads to better and more cost-efficient model as calibration metric, (b) \passk can understate or overstate differences in model capability under sequential feedback.
    (c) \seqk often exposes superior Pareto frontier at deployment than that of \passk 
    and (d) how \seqk tracks a different capability from \passk (Section~\ref{subsec:modelselection}).}
     }

    \item \textbf{Uncovering Revision Patterns:} {Beyond reporting \seqk, 
    we further analyze the way in which models revise their attempts: (a) models can differ in when they gain, with some gaining most on the first retry and others on later attempts,
    (b) revisions do not always address the feedback, and some models resubmit the same attempt unchanged,
    and (c) even revisions that address all of the feedback can fail by breaking requirements that passed before (Section~\ref{subsec:heterogeneity}).}

    \item \textbf{Diagnosing the Feedback Loop:} {\seqk also helps diagnose the feedback itself.
    We show (a) the usefulness of informative feedback depends on both the benchmark and the model capability,
    (b) critiques written with the grading rubric tend to help models repair far more failed requirements, 
    and (c) summarizing the previous attempts can lower success rate, so how the history is kept should be reported along with \seqk (Section~\ref{subsec:informativeness}).}
\end{enumerate}

\section{Related Work}
\paragraph{Refinement from feedback.}
A line of related work considers having a model reasoning and revising in response to critique or execution feedback as methods to improve agent performance \citep{madaan2023selfrefine, shinn2023reflexion, gou2024critic}. 
Reflexion \citep{shinn2023reflexion} is the closet to our setting by considering agent's reasoning on top of binary pass/fail signal or results on self-generated test cases. 
In contrast, our work centers on \seqk as an evaluation metric, and as a result, considers situation where the feedback is provided under privileged information about the correct answer or grading criteria. 
This reflects the common setting where users of an agent provides information on how its attempts were inadequate. 
Other recent work evaluate this ability by introducing novel tasks. 
RefineBench \citep{lee2025refinebench} introduces a new set of tasks with checklist-based evaluation, and finds that self-refinement yields little improvement while targeted feedback drives frontier models close to saturation. 
ICPC-Eval's Refine@K \citep{xu2025icpceval} measures revision from execution feedback in competitive programming. 
Instead of constructing new tasks, we demonstrate that on existing benchmarks and settings the community already considers, the \seqk evaluation protocol already reveals insights on model performance rankings and novel cost-performance Pareto frontiers. 
These insights demonstrate how \seqk is an important metric to report alongside \passk on benchmarks or settings where iterative refinement is expected in practice. 

\paragraph{Multi-turn and agentic evaluation.}
Agent benchmarks such as SWE-bench, WebArena, and $\tau$-bench \citep{jimenez2024swebench, zhou2023webarena, yao2024taubench} and multi-turn evaluations such as MINT \citep{wang2024mint} score models over long interactions with tools and environments. 
In these settings, success depends jointly on planning, tool use, state tracking, and the use of feedback.
\seqk isolates the last of these, where each piece of feedback provides information on the entire attempt.
Hence, differences in \seqk reflect how models revise their attempts and not how they navigate an environment.

\section{\seqk Evaluation Protocol}
\label{sec:metric}

\revised{Both \passk and \seqk evaluate whether a model {succeeds at least once} within an attempt budget of $K$, but they differ in whether candidate draws are independent or generated through iterative refinement from feedback. 
Independent sampling generates each candidate from a static task prompt, whereas sequential evaluation conditions each revision on preceding candidate responses and {feedback}. {We mathematically formulate these metrics as follows.}}

Let $\mathcal{D} = \{x_i\}_{i=1}^{N}$ denote a benchmark of $N$ tasks, where $x_i$ is the initial prompt for task $i$. Let $\mathcal{Y}$ denote the candidate solution space. Each task $x_i$ is paired with a verifier $V_i: \mathcal{Y} \to \{0,1\}$ indicating whether a candidate $y \in \mathcal{Y}$ passes the success requirement. {The verifier may be an LLM if the evaluation protocol of the benchmark requires an LLM grader.} A model with parameters $\theta$ induces a conditional decoding distribution $\pi_\theta(\cdot \mid c)$ over $\mathcal{Y}$ given an input context $c$. {We call this model the actor, to distinguish it from the models that write feedback (critics) or grade attempts (graders).}

\paragraph{\passk.}
Under independent sampling, context is restricted strictly to the original task prompt ($c = x_i$). For each task $x_i$, the model independently draws $K$ candidate solutions $Y_{i,1}, \dots, Y_{i,K} \sim \pi_\theta(\cdot \mid x_i)$. A task succeeds if at least one candidate passes the verifier $V_i$. The \passk metric computes the expected success rate across the benchmark:
\begin{equation}
\mathrm{Pass@}K = \frac{1}{N} \sum_{i=1}^{N} \mathbb{E}_{Y_{i,1},\dots,Y_{i,K} \sim \pi_\theta(\cdot \mid x_i)} \left[ \max_{1 \le t \le K} V_i(Y_{i,t}) \right].
\label{eq:passk}
\end{equation}
Because $V_i(y) \in \{0,1\}$, the maximum over attempt steps $t$ evaluates to $1$ if any attempt succeeds and $0$ otherwise.

\paragraph{\seqk.}
Under sequential evaluation, attempts are generated iteratively, allowing the prompt context to expand with past candidate responses and environmental feedback. Let $\mathcal{F}$ denote the space of feedback messages. A feedback mode $m$ specifies what feedback message the actor receives after a failed attempt (Section~\ref{sec:setup}). The interaction history before attempt $t$ is $H_{i,t-1} = \big((Y_{i,1}, F_{i,1}), \dots, (Y_{i,t-1}, F_{i,t-1})\big)$, with $H_{i,0} = \emptyset$.

At attempt $t \in \{1, \dots, K\}$, the model samples $Y_{i,t} \sim \pi_\theta(\cdot \mid x_i, H_{i,t-1})$. If the attempt fails ($V_i(Y_{i,t}) = 0$), the environment returns feedback $F_{i,t} \in \mathcal{F}$ according to the feedback mode $m$, which is appended to $H_{i,t-1}$ for the next turn. 
If the attempt succeeds ($V_i(Y_{i,t}) = 1$), interaction terminates early. 
We call an attempt with $t \ge 2$, which is conditioned on earlier attempts and feedback, a revision or a revised attempt. 
Let $\tau_i$ denote the resulting sequential trajectory, and let $P_{\theta,m,i}^{(K)}(\cdot \mid x_i)$ denote the joint distribution over trajectories induced by the model, feedback mode $m$, and budget $K$. The \seqk metric evaluates the expected trajectory success rate:
\begin{equation}
\mathrm{Seq@}K = \frac{1}{N} \sum_{i=1}^{N} \mathbb{E}_{\tau_i \sim P_{\theta,m,i}^{(K)}(\cdot \mid x_i)} \left[ \max_{1 \le t \le K} V_i(Y_{i,t}) \right].
\label{eq:seqk}
\end{equation}
When a trajectory halts early upon success, the maximum is evaluated over generated attempts within budget $K$. Unlike \passk, successive candidate attempts in \seqk are not independent draws from a static prompt; each revision depends directly on previous attempts and feedback.

\paragraph{Feedback Modes and Protocol Dependencies.}
The \seqk metric forms a family of feedback-conditioned evaluations indexed by the feedback mode $m$. A numerical \seqk value is well-defined only after specifying feedback mode $m$ alongside the benchmark task split, verifier, prompt builder, decoding hyperparameters, and attempt budget.

\section{Experimental Design}
\label{sec:setup}

\revised{{We evaluate $7$ benchmark variants spanning instruction following, rubric-graded generation, and executable code synthesis.} Table~\ref{tab:study-coverage} outlines these benchmarks alongside their experimental configurations and the Results sections that use them.
Each empirical comparison uses actors, task splits, and protocol settings drawn from this central collection, with matching criteria and remaining protocol differences specified in the appendix.
Detailed definitions for all benchmarks, success criteria, task screening procedures, and decoding parameters are provided in Appendix~\ref{app:setup}.}

\begin{table}[ht]
\centering\small\revisioncolor
\caption{\revised{\textbf{Experimental coverage across evaluation cohorts.} Overview of completed run records associated with our analyses across the 7 evaluated benchmark variants. 
This table summarizes the actor coverage, run counts, and the Results sections each benchmark is associated with. 
The Actors column counts distinct models evaluated across benchmarks.
}}
\label{tab:study-coverage}
\begin{tabularx}{\linewidth}{@{}>{\raggedright\arraybackslash}p{.40\linewidth}rr>{\raggedright\arraybackslash}X@{}}
\toprule
Benchmark & Actors & Runs & Used in \\
\midrule
AdvancedIF \citep{he2025advancedif} & 10 & 61 & Sections~\ref{subsec:modelselection}, \ref{subsec:heterogeneity}, \ref{subsec:informativeness} \\
CLBench DKR \citep{dou2026cl} & 8 & 80 & Sections~\ref{subsec:modelselection}, \ref{subsec:heterogeneity}, \ref{subsec:informativeness} \\
CLBench RSA \citep{dou2026cl} & 3 & 8 & Section~\ref{subsec:modelselection} \\
HealthBench \citep{arora2025healthbench} & 7 & 67 & Sections~\ref{subsec:modelselection}, \ref{subsec:informativeness} \\
ResearchRubrics \citep{sharma2025researchrubrics} & 8 & 35 & Sections~\ref{subsec:modelselection}, \ref{subsec:informativeness} \\
MBPP Pro \citep{yu2025humanevalpro} & 4 & 112 & Sections~\ref{subsec:modelselection}, \ref{subsec:heterogeneity}, \ref{subsec:informativeness} \\
HumanEval Pro \citep{yu2025humanevalpro} & 4 & 108 & Sections~\ref{subsec:modelselection}, \ref{subsec:heterogeneity}, \ref{subsec:informativeness} \\
\midrule
Total & 16 & 471 & \\
\bottomrule
\end{tabularx}
\end{table}

\paragraph{\revised{Feedback modes.}}
\revised{We compare independent parallel sampling (\passk) against three distinct sequential feedback modes, categorized by the signal provided to the actor following a failed attempt:}
\begin{itemize}[leftmargin=*,nosep,label={$\bullet$}]
    \item \textbf{Binary Feedback:} Provides only an outcome pass/fail verdict. This mode evaluates whether knowing that a candidate failed provides sufficient leverage to guide subsequent revision.
    \item \revised{\textbf{Raw Verifier Feedback:} Returns direct environment or evaluator diagnostics—including execution stack traces, failed test assertions, or rubric item breakdowns. This mode tests whether models can interpret and act upon evaluator diagnostics.}
    \item \revised{\textbf{LLM Critic Feedback:} Uses a language model to generate targeted critique from the candidate response and the information supplied to the critic. The critic role can be fulfilled by the actor model itself or another language model. We systematically vary the critic's context, including whether it has explicit access to underlying grading rubrics.}
\end{itemize}
\revised{Feedback mode specifies the information boundary maintained during revision rounds.  For example, raw test diagnostics can expose test failures and expected outputs, whereas rubric-aware critics use evaluation rubrics to generate guidance for the actor.

{Having established the benchmark coverage, feedback channels, and evaluation protocols, we next present our empirical findings on how \seqk reshapes model selection, exposes {revision dynamics}, and quantifies feedback utility across these systems.}}

\section{Results}
\label{sec:results}

\revised{We structure our empirical investigation around three core questions: (1) Does evaluating models under sequential feedback alter model selection and cost--performance trade-offs? (2) How do {revisions} unfold across successive feedback turns? (3) What properties of the feedback channel determine whether an attempt succeeds or stalls?}

\subsection{Model-selection under \seqk}
\label{subsec:modelselection}\label{subsec:selection}

\subsubsection{More performant and efficient model selection with \seqk}
\begin{figure}[!ht]
\centering
\begin{subfigure}[t]{0.49\linewidth}
\centering
\includegraphics[width=\linewidth]{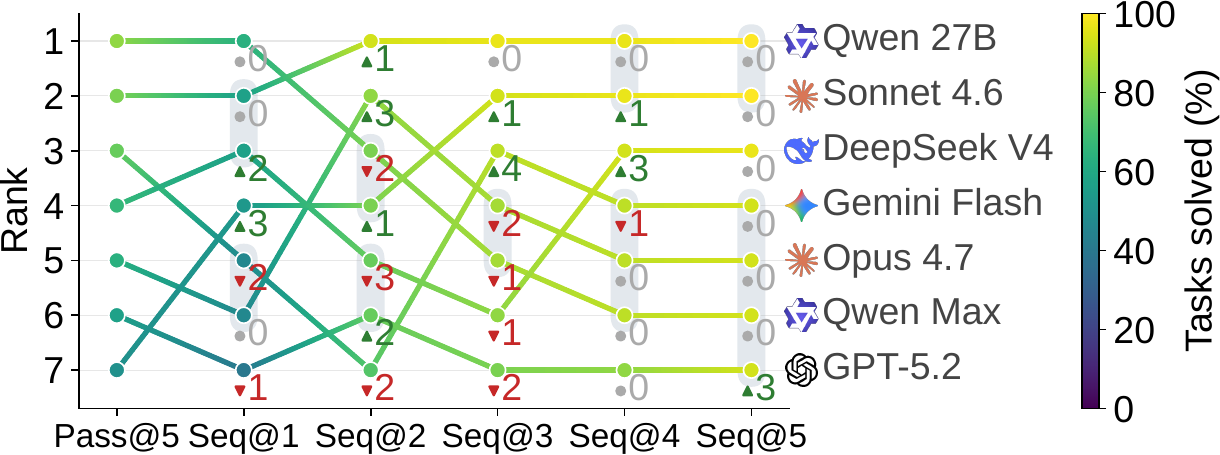}
\caption{AdvancedIF}
\label{fig:rank-change-advancedif}
\end{subfigure}%
\hfill
\begin{subfigure}[t]{0.49\linewidth}
\centering
\includegraphics[width=\linewidth]{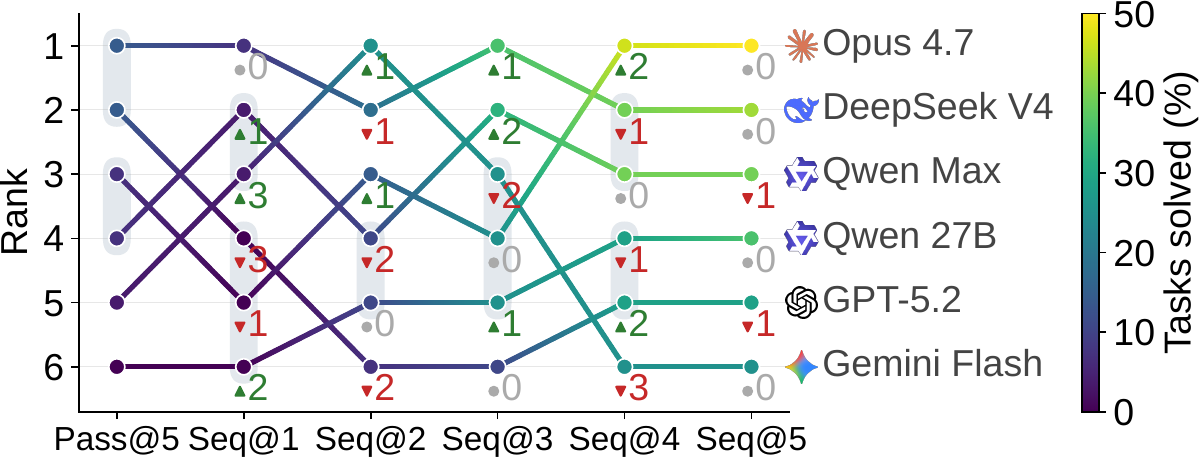}
\caption{CLBench DKR}
\label{fig:rank-change-clbench-dkr}
\end{subfigure}
\caption{\revised{\textbf{Feedback changes model ordering.} (a) AdvancedIF: $7$ actors, $30$ shared tasks, raw feedback. (b) DKR: $6$ actors, $28$ shared tasks, actor-specific critic configurations. All use $T=0.7$. Rank trajectories run from \passfive through Seq@1--5, colored by success rate. The number below each point is the rank change from the previous column. Shaded groups are tied and share the group's best rank; order within a tie is arbitrary, and a move within a tie is no rank change. These are descriptive orderings. Full scores and configurations: Appendix~\ref{app:gain-correlation}.}}
\label{fig:advancedif-separation}
\end{figure}

\revised{A model's ranking under static, single-attempt evaluation (\passfive) can diverge significantly from its ranking under multi-turn sequential evaluation (\seqfive). As shown in Figure~\ref{fig:advancedif-separation}, Sonnet climbs from the bottom rank under independent sampling on AdvancedIF to the top tied group after sequential feedback, with similar order inversions occurring on CLBench DKR. Given this disparity, the practical question is whether evaluating the intended feedback loop helps us choose a better model with limited calibration spending.}

\revised{{\textbf{Selection test.} We simulate a practitioner who picks an actor on a few calibration tasks and then deploys it with sequential feedback on new tasks. We compare picking by \passfive with picking by \seqk at a shorter depth ($K=3$ or $4$).
Since each sequential attempt carries the earlier attempts and feedback in its context, \seqk can be more costly.
We therefore also record the calibration cost.
Each candidate actor runs three times on the same calibration tasks at $T=0.7$, and we average its success over the three repetitions.
We pick the actor with the highest score in calibration tasks (breaking ties uniformly at random), and evaluate it under raw-verifier-feedback (with details on failed tests) \seqfive on held-out tasks.
\emph{Selection loss} is the main metric we use to evaluate the performance of selection process, which is the held-out score gap between the picked actor and the best candidate.}}

\begin{figure}[!ht]
\centering
\includegraphics[width=\linewidth]{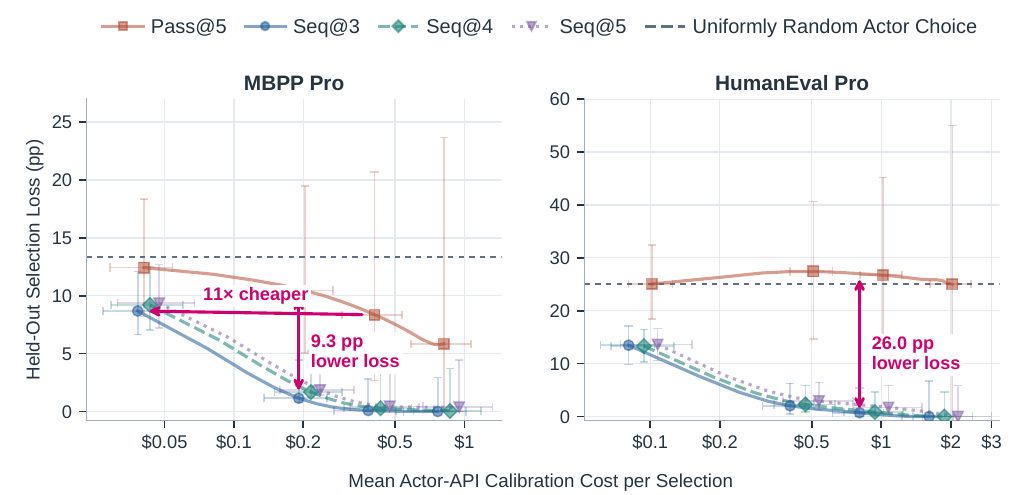}
\caption{\revised{\textbf{Short sequential calibration picks better actors for sequential deployment.} Selection loss and calibration cost for $n=1$--$20$ calibration tasks ($T=0.7$). The picked actor is evaluated under \seqfive with raw verifier feedback (containing details on failed tests). Markers show $n \in \{1, 5, 10, 20\}$, and Table~\ref{tab:budgeted-selection-main} reports $n=10$. On MBPP Pro, \seqthree calibration on $1$ task reaches about the same selection loss as \passfive on $10$ tasks at $11\times$ lower cost. Vertical arrows mark the largest gap between \seqthree and \passfive among the plotted task counts. Error bars are $95\%$ bootstrap intervals. The dashed line is a uniformly random actor choice baseline, which incurs no calibration cost.
}}
\label{fig:selection-main}

\end{figure}

\revised{{We evenly split the $45$ MBPP Pro tasks and the $40$ HumanEval Pro tasks (Appendix~\ref{app:interface-audit}) into two fixed groups per benchmark, group A and B.
We calibrate on A and evaluate on B, and vice versa.
For a given calibration and evaluation group, we repeatedly draw $n$ tasks for each $n \in [20]$  from the calibration group, with the evaluation task set held fixed across draws and calibration sizes.
Figure~\ref{fig:selection-main} reports selection loss and calibration cost averaged over draws and both directions, with each direction weighted by the size of its evaluation group. Table~\ref{tab:budgeted-selection-main} reports $n=10$. {Confidence intervals come from bootstrap (Appendix~\ref{app:budgeted-selection}).}}}

\begin{table}[!ht]
\centering\small\revisioncolor
\setlength{\tabcolsep}{3.5pt}
\caption{\revised{{\textbf{Selection with 10 calibration tasks.} The mean calibration cost per selection and the held-out \seqfive of the picked actor, with $T=0.7$ on $45$ MBPP Pro and $40$ HumanEval Pro tasks. \seqk calibration uses $K=3$ or $4$.
Note that both \seqk and \passk stop at the first success without incurring additional cost.
Gain shows the improvement from better picked actor from calibrating with \seqfive compared to that of \passfive.
}}
}
\label{tab:budgeted-selection-main}

\begin{tabular}{lcrrrrr}
\toprule
& & \multicolumn{2}{c}{Mean cost (USD)} & \multicolumn{3}{c}{Chosen actor's Seq@5 (\%)} \\
\cmidrule(lr){3-4}\cmidrule(lr){5-7}
Benchmark & $K$ & Pass@5 & Seq@$K$ & Pass@5 pick & Seq@$K$ pick & Gain (pp) \\
\midrule
MBPP Pro & 3 & 0.407 & 0.383 & 90.2 & 98.4 & $+8.3$ \\
MBPP Pro & 4 & 0.407 & 0.432 & 90.2 & 98.2 & $+8.1$ \\
HumanEval Pro & 3 & 1.015 & 0.805 & 70.8 & 96.8 & $+26.0$ \\
HumanEval Pro & 4 & 1.015 & 0.937 & 70.8 & 96.7 & $+25.9$ \\
\bottomrule
\end{tabular}
\end{table}

\revised{{\textbf{Finding.} Calibrating with \seqthree or \seqfour {picks significantly better actors for sequential deployment than calibrating with \passfive, at similar or lower cost} (Figure~\ref{fig:selection-main} and Table~\ref{tab:budgeted-selection-main}). The gain is largest on HumanEval Pro, where actors that look similar under \passfive differ widely under \seqfive (Table~\ref{tab:separation-main}).}}

\subsubsection{\passk can overstate or understate performance gaps}
The cost of selecting a worse actors
depends on how far apart the candidate actors are in performance (Table~\ref{tab:separation-main}).
On AdvancedIF, the gap between the best and worst actor is $33.3$ points under \passfive but only $6.7$ points under \seqfive with raw verifier feedback.
This is one example where, as long as proper feedback is provided, the performance gap between actors is not as large.
HumanEval Pro is a contrasting example, where actors differ even more with feedback that gives details on failed tests ($10.8 \rightarrow 51.7$ points from \passfive to \seqfive).
Here, actors differ in how well they use the feedback in their next attempts, and the tasks are ones where this ability matters.
The average gap over all model pairs moves in the same directions, suggesting that this is a prevalent pattern in each benchmark.
On MBPP Pro, the gap also widens.
These cases show that \passk can significantly understate or overstate the differences between actors, so where sequential interaction is expected, \seqk should be considered instead.

\begin{table}[!ht]
\centering\small\revisioncolor
\caption{\textbf{\passfive can overstate or understate the performance gap between actors.} The gap between the best and worst actor and the average gap over all model pairs, with $T=0.7$.
 \seqfive uses raw verifier feedback on AdvancedIF and details on failed tests on MBPP Pro and HumanEval Pro. Change is the gap under \seqfive minus the gap under \passfive.
 Details are provided in Appendix~\ref{app:separation-uncertainty}.}
\label{tab:separation-main}
\begin{tabular}{lrrr}
\toprule
& \multicolumn{2}{c}{Best--worst model gap} & All-pairs gap \\
Benchmark & Pass@5 $\to$ Seq@5 & Change & Change \\
\midrule
AdvancedIF & $33.3 \to 6.7$ & $-26.7$ & $-11.7$ \\
MBPP Pro & $18.5 \to 31.1$ & $+12.6$ & $+8.6$ \\
HumanEval Pro & $10.8 \to 51.7$ & $+40.8$ & $+25.4$ \\
\bottomrule
\end{tabular}
\end{table}

\subsubsection{\seqk reveals better Pareto-optimal choices}
\label{subsec:resources}
\begin{figure}[!ht]
\centering
\begin{subfigure}[t]{0.49\linewidth}
\centering
\includegraphics[width=\linewidth]{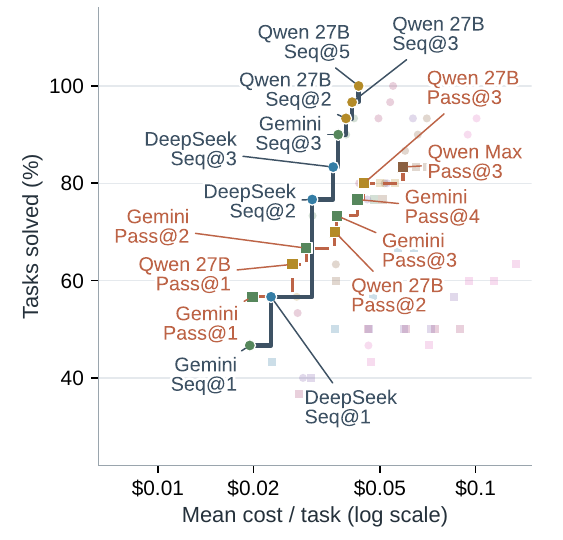}
\caption{AdvancedIF}
\label{fig:cost-frontier-advancedif}
\end{subfigure}%
\hfill
\begin{subfigure}[t]{0.49\linewidth}
\centering
\includegraphics[width=\linewidth]{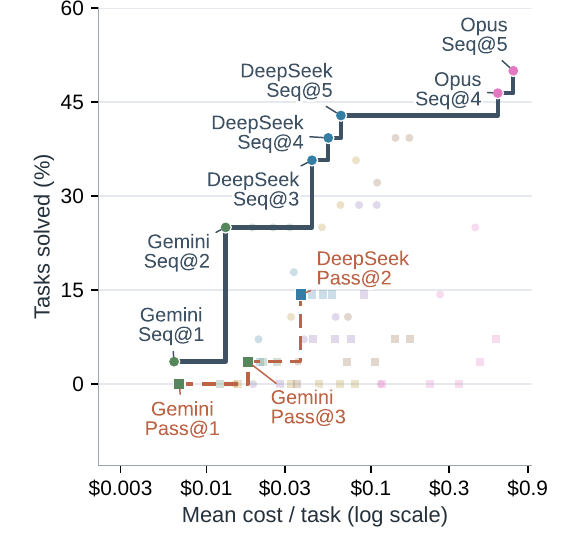}
\caption{CLBench DKR}
\label{fig:cost-frontier-clbench-dkr}
\end{subfigure}

\vspace{2pt}
\includegraphics[width=\linewidth]{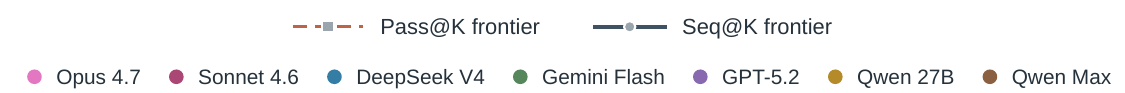}
\caption{\revised{{\textbf{Sequential feedback expands cost--accuracy frontiers.}
Each frontier shows the highest success reached for a given mean cost per task, for $K=1$ to $5$ with $T=0.7$. Both \seqk and \passk stop incurring cost at the first success.
Labeled markers are on the frontier and show the actor and $K$, and faint markers are not on the frontier.
{\seqk tends to reveal a better Pareto frontier in both}
(a) AdvancedIF across $7$ actors, {$30$ shared tasks,} and raw verifier feedback and (b) DKR across $6$ actors, $28$ shared tasks.
}}}
\label{fig:cost-comparison}

\end{figure}

\revised{{In practice, one could deploy an agent for $K$ parallel attempts as represented by \passk.
\seqk represents the alternative of revising with feedback for up to a predetermined $K$ attempts.
While each sequential attempt carries more context from previous sessions, they also tend to have earlier successes from incorporating feedback.
This motivates us to explore the cost-performance Pareto frontier of both protocols.

Figure~\ref{fig:cost-comparison} shows the Pareto frontier of each protocol over two benchmarks, all actors, and $K=1$ to $5$.
\seqk often reveals a better Pareto frontier than \passk, with higher success for the same cost.
The difference is largest on DKR, where independent sampling solves at most about $14\%$ of tasks, while sequential feedback reaches $50\%$ (Opus 4.7 at $K=5$).
}}

\subsubsection{\seqk tracks a distinct capability signal from \passk}
\begin{figure}[!ht]
\centering
\includegraphics[width=\linewidth]{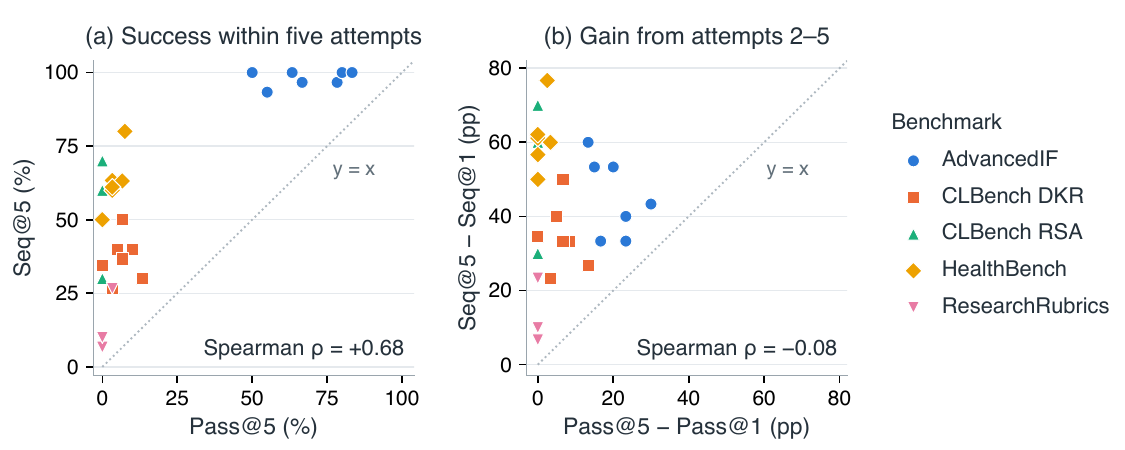}
\caption{\revised{\textbf{Final success correlates across protocols, but gains from retries do not.} Each point is one model--benchmark pair with $K=5$. \seqfive uses the best non-binary feedback configuration for each pair. (a) \seqfive against \passfive. (b) The gain from attempts 2 to 5 under each protocol. Details are in Appendix~\ref{app:gain-correlation}.}}
\label{fig:gain-comparison}
\end{figure}

\revised{The previous results show the operational implications of \seqk for choosing models.
We complement them with a more direct study of how \seqk tracks a different capability from \passk.
At first glance, the two metrics are somewhat correlated.
Over $26$ model--benchmark pairs, \passfive and \seqfive have Spearman correlation $+0.68$ (Figure~\panelref{fig:gain-comparison}{a}).
This is because the two protocols have approximately the same success rate on the first attempt, which reflects the actor's one-shot capability on the task.
To control for this, we measure instead how much attempts 2 to 5 add under each protocol:}
\[
\revised{\Delta\passk=\passk-\passone,\qquad
\Delta\seqk=\seqk-\seqone.}
\]
\revised{With the first attempt removed, the correlation drops to $-0.08$ (Figure~\panelref{fig:gain-comparison}{b}).
That is, how much a model gains from more independent samples says little about how much it gains from feedback.
Hence, the iterative refinement capability that \seqk evaluation aims to track is distinct from what \seqk captures.
}

\subsection{{Seq@K reveals when and how revisions succeed}}
\label{subsec:heterogeneity}

The numerical results on success rates earlier do not show when revisions succeed or how an actor responds to feedback.
But the rate at which {agents'} revisions succeed and how well the revisions address the feedback directly impacts the usability of agentic systems.
In this section, we examine when revised attempts yield gains and failure modes in agents' revision patterns.

\subsubsection{Revision timing distinguishes front-loaded from distributed gains}
\revised{Models with similar \seqfive can follow distinct revision trajectories across turns. This profile matters when deployment limits the number of revision turns. To characterize this behavior, we analyze how performance improvements distribute across revision turns. For any model exhibiting net improvement ($\Delta\seqfive > 0$), we compute the Early-Gain Share (EGS) and Last-Gain Share (LGS):}
\[
\revised{\mathrm{EGS}=\frac{\seqtwo-\seqone}{\Delta\seqfive},\qquad
\mathrm{LGS}=\frac{\seqfive-\seqfour}{\Delta\seqfive}.}
\]
\revised{The early-gain share (EGS) is the fraction of improvement obtained on the first retry. The last-step share (LGS) is the fraction obtained on the final allowed attempt.}

\revised{We evaluate these metrics using rubric-guided DKR setups across two distinct critic styles: \emph{directive} feedback (providing explicit actionable revision instructions) and \emph{socratic} feedback (providing targeted probing questions). Table~\ref{tab:clbench-dkr-feedback-gain} summarizes the distribution of gains across turns for these configurations.
In the selected configurations, Gemini 3 Flash obtains most of its gain on the first retry with the highest EGS, which may be preferable for settings where earlier revisions are preferred.
On the other hand, a non-zero LGS like that of Opus 4.7 or {DeepSeek} V4 Pro indicates potential gains from more repeated attempts.
}

\begin{table}[!ht]
\centering\small\revisioncolor
\caption{\revised{\textbf{Some configurations revise earlier than others.} 
On selected CLBench DKR configurations and $30$ tasks. 
EGS and LGS breakdown when each actor successfully revise their attempts, with higher EGS potentially more suitable where earlier revisions are preferred, and non-zero LGS indicating the need for more sequential attempts allocated. 
More details are provided in Appendix~\ref{app:clbench-dkr-attempt-deltas}.
}
}
\label{tab:clbench-dkr-feedback-gain}
\begin{tabular}{llrrrr}
\toprule
Actor & Feedback / critic & Seq@1 & Seq@5 & EGS & LGS \\
\midrule
Opus 4.7 & Directive / actor itself & 0.000 & 0.500 & 0.33 & 0.07 \\
DeepSeek V4 Pro & Socratic / GPT-5.2 & 0.067 & 0.400 & 0.30 & 0.10 \\
Gemini 3 Flash & Socratic / GPT-5.2 & 0.033 & 0.267 & 0.86 & 0.00 \\
GPT-5.2 & Directive / actor itself & 0.033 & 0.300 & 0.25 & 0.00 \\
Qwen3.6 Max & Socratic / GPT-5.2 & 0.033 & 0.367 & 0.20 & 0.00 \\
\bottomrule
\end{tabular}
\end{table}

\subsubsection{Revisions do not always address the feedback}\label{subsubsec:revision-feedback}
\begin{figure}[!ht]
\centering
\includegraphics[width=\linewidth]{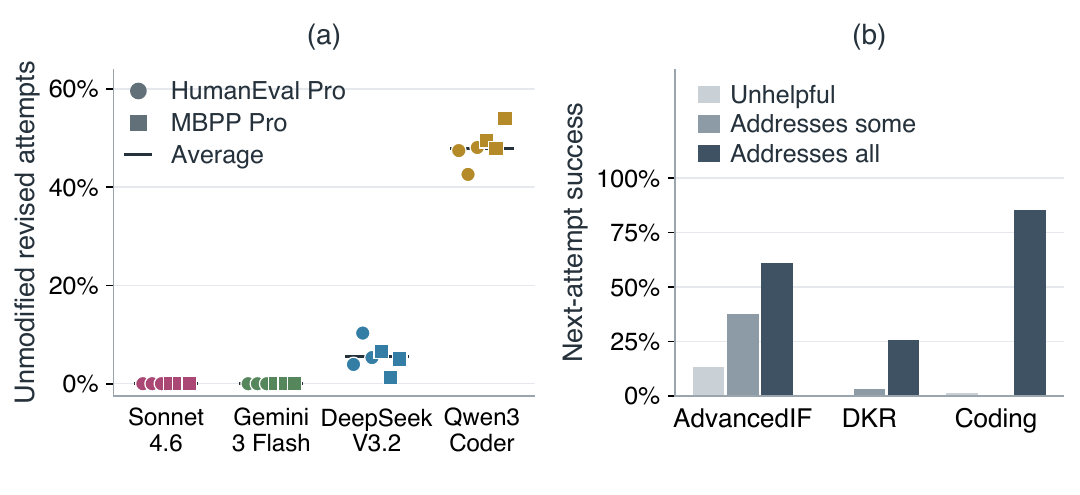}
\caption{\textbf{Revisions do not always address the feedback.} (a) The fraction of revised attempts in coding that are exactly the same as the previous attempt, for four models on HumanEval Pro and MBPP Pro.
Each point is one of three independent runs on each benchmark, and the line shows the average over all six runs.
{DeepSeek} and Qwen models appear more likely to submit revisions identical to the previous failed attempts.
(b) The success rate of the next attempt, grouped by whether the revision is unhelpful, addresses some, or addresses all of the feedback.
The issue of breaking previously passing criteria is particularly pronounced in the DKR and {AdvancedIF} benchmarks, where success requires meeting multiple criteria, and less so the case in coding benchmarks.
}
\label{fig:repair-trace}\label{tab:qualitative-transitions}
\end{figure}

We next examine whether provided feedback are addressed in agents' next attempts.
Some unhelpful revisions are easy to identify since the next attempt entails the same answer as before with no changes.
In coding, we find such examples with Qwen3 Coder Flash and DeepSeek V3.2:
$48\%$ and $6\%$ of revised attempts respectively are exactly the same attempt as before without modifications, whereas Sonnet 4.6 and Gemini 3 Flash never submit an unmodified revised attempt (Figure~\panelref{fig:repair-trace}{a}).

Other unhelpful revisions update the attempts but nonetheless fail to address the feedback.
Unlike the aforementioned failure mode that can be programmatically identified, finding cases where revisions took place but are ineffective takes manual inspection.
Below, we show 2 of the 14 examples we identify (Examples~\ref{card:sidehustle-gpt} and~\ref{card:spartan}) {and more in Appendix~\ref{app:sec53-diagnostic-alignment}}.

\begin{revexample}[label=card:sidehustle-gpt]{AdvancedIF}{GPT-5.2}
\revfield{Task}{The assistant is a side-hustle advisor whose instructions require it to decline off-topic questions. The user asks whether AI makes painting irrelevant.}
\revfield{Feedback}{Attempt 1 answers the off-topic question instead of declining it.
}
\revfield{Revision}{Each revised attempt keeps arguing that AI does not make painting irrelevant and only swaps the side hustle it recommends: live event painting, then pet portraits, then painting commissions.}
\revfield{Why Unhelpful}{The revisions change content the feedback never questioned, so the same criterion fails on all five sequential attempts.
On the other hand, when Gemini 3 Flash is given the same kind of feedback on this task, it adds the refusal in its third attempt and passes in its fourth.}
\end{revexample}

\begin{revexample}[label=card:spartan]{CLBench DKR}{DeepSeek V4 Pro}
\revfield{Task}{The assistant is a strength-training coach for women aged 18--70. A 20-year-old user who plays on a men's netball team asks for a chest training plan.}
\revfield{Feedback}{Attempt 1 refuses the request as expected, but the rubric requires the refusal to rest explicitly on the user's gender.
}
\revfield{Revision}{Each revised attempt rewords the refusal but still justifies it by the user's netball team.}
\revfield{Why Unhelpful}{The refusal never explicitly states the user's gender as the reason, so the criterion fails on all five attempts.}
\end{revexample}

Since manually going through all revisions is not feasible, we employ a similar approach to GEPA~\citep{agrawal2026gepa}: we use the 14 examples to develop a prompt, and use Claude Opus 5.5 to annotate all tasks in the four benchmarks (AdvancedIF, CLBench DKR, HumanEval Pro, and MBPP Pro) on attempts that contain revisions.
For each feedback, the annotator judges whether the revised attempt satisfies it.
A revision is \emph{unhelpful} if it satisfies none of the remaining requests, \emph{addresses some} of the feedback if it fully or partly satisfies at least one, and \emph{addresses all} of the feedback if it fully satisfies every one. Across $842$ revisions, we find that $186$ ($22\%$) are unhelpful, $289$ ($34\%$) address some, and $367$ ($44\%$) address all of the feedback (Appendix~\ref{app:repair-generalization}).

In Figure~\panelref{fig:repair-trace}{b}, we further show how unhelpful revisions impact task successes in the sequential setting.
First, as expected, unhelpful revisions almost never succeed in the next attempt.
Success rises when some of the feedback are deemed partially addressed.
However, addressing all of the feedback is not enough and many such revisions still fail;
this is because the revisions can break a criterion that passed before as it addresses the current feedback.
This failure mode is {particularly} pronounced in the DKR and {AdvancedIF} benchmarks, where success requires meeting multiple criteria, and less so the case in coding benchmarks.

\subsection{\revised{Seq@K diagnoses the feedback loop}}
\label{subsec:informativeness}
{
So far, we examined how agents perform under \seqk and respond to feedback.
In this section, we look at the feedback itself: what information it contains, whether the critic knows what is being graded, and how much of the previous attempts the agent sees.}
\subsubsection{Effectiveness of binary feedback depends on the benchmark}

{Binary feedback only tells the agent that its attempt failed, without saying which requirements are not met.
In DKR, binary feedback is largely unhelpful.
To measure this, we take the requirements that the first attempt fails and measure the fraction of them fixed in the best later attempt (the attempt that satisfies the most requirements, taking the latest one if tied). We call this fraction \emph{criterion repair}.
In Table~\ref{tab:feedback-pooled}, we compare independent retries, binary feedback, and critiques written with and without access to the grading rubric.
Binary feedback and critiques without the rubric repair about as many criteria as independent retries, whereas critiques with the rubric repair many more.
Binary feedback is similarly unhelpful for DeepSeek V3.2 and Qwen3 Coder Flash in coding, where it changes \seqfive by at most $3.3$ points from \passfive (Appendix~\ref{app:coding-results}).}

{On the other hand, binary feedback is helpful for Sonnet 4.6 in coding, since Sonnet can reason on its own about why its code failed when it was run.
Compared to \passfive, \seqfive with binary feedback is $17$ to $26$ points higher for Sonnet on HumanEval Pro and MBPP Pro (Appendix~\ref{app:coding-results}), whereas in DKR it is only $1$ point higher on average across six models.}
Overall, this shows that effectiveness of different feedback type is dependent on both benchmark and model-capability.
Therefore, in designing feedback harnesses around models, practitioners may benefit more from highlighting criteria that agents would otherwise not come up on their own from reasoning.
\begin{table}[!ht]
\centering\small\revisioncolor
\caption{{\textbf{In DKR, feedback is more useful when the critic knows what is being graded.} The fraction of requirements failed in the first attempt of DKR that are fixed in a later attempt, pooled over five models with $30$ tasks each. Binary feedback and critiques without the rubric repair about as many requirements as independent retries. Details are in Appendix~\ref{app:sec53-binary}.}}
\label{tab:feedback-pooled}%
\begin{tabular}{llr}
\toprule
Condition & Actor receives & Failed criteria repaired (\%) \\
\midrule
Independent sampling & No earlier answer or feedback & 36.5 \\
Binary feedback & Earlier answers + failure verdict & 34.4 \\
Rubric-blind critic & Critique without grading criteria & 36.8 \\
Rubric-aware critic & Critique informed by grading criteria & \textbf{73.8} \\
\bottomrule
\end{tabular}

\end{table}

\subsubsection{Summarizing the history can lower sequential success}
\label{subsec:history}
{In \seqk, the prompt for each attempt includes the latest feedback and the previous attempts.
When we replace the previous attempts with a summary, performance changes little on AdvancedIF and coding but drops on the other benchmarks, most consistently on HealthBench (Table~\ref{tab:loop-controls}).
How the history is kept is therefore part of the protocol and should be reported along with \seqk.
Details are provided in Appendix~\ref{app:protocol-ablations}.
}
\begin{table}[!ht]
\centering\small\revisioncolor
\caption{\textbf{Summarizing the history can lower sequential success.} Here, each unit is a pair of trajectories on the same task where one has previous attempts summarized, and one keeps the full history.
The two trajectories otherwise use the same model, feedback, grader, and attempt budget.
The table shows the average change in success rate from summarization over these pairs.
{Performance changes little on AdvancedIF and coding but drops on the other benchmarks.}}
\label{tab:loop-controls}
\begin{tabular}{lrr}
\toprule
Benchmark & Trajectory pairs & Change (pp) \\
\midrule
AdvancedIF & 200 & $-0.5$ \\
MBPP Pro & 800 & $+1.8$ \\
HumanEval Pro & 800 & $+0.5$ \\
CLBench DKR & 180 & $-6.1$ \\
ResearchRubrics & 180 & $-7.2$ \\
HealthBench & 360 & $-14.4$ \\
\bottomrule
\end{tabular}
\end{table}

\section{Discussion and Conclusion}
\label{sec:conclusion}

{We introduced \seqk, which measures whether a model solves a task within $K$ attempts when it receives feedback after each failed attempt.
Model rankings, gaps between actors, and cost-performance frontiers can all differ between \seqk and \passk, so choosing a model by \passk can pick the wrong one for an agent that revises from feedback.
\seqk trajectories also show when models gain and how revisions fail, for example by ignoring the feedback or by breaking requirements that passed before.
Finally, \seqk helps diagnose the feedback loop itself. 
We find how much feedback helps depends on what it tells the agent, whether the critic knows what is being graded, and how the history is kept.}
Practitioners building agentic systems around a feedback loop should choose models with \seqk under the feedback the agent will receive, since \passk may favor a model that does not use that feedback well.

\bibliography{references}
\bibliographystyle{conference}

\appendix
\newpage
\section{Benchmark and Setup Details}
\label{app:setup}\label{app:detailed-setup}

Table~\ref{tab:detailed-benchmark-setup} lists the benchmarks and when an attempt counts as a success. The rest of this appendix explains how feedback, history, and grading are set up (Appendix~\ref{app:feedback-history-grading}), gives the exact prompts and feedback used in coding (Appendix~\ref{app:reference-protocol}), and lists the coding tasks we exclude (Appendix~\ref{app:interface-audit}).

\begin{table}[h]
\centering\small
\begin{tabularx}{\linewidth}{>{\raggedright\arraybackslash}p{0.20\linewidth}>{\raggedright\arraybackslash}X>{\raggedright\arraybackslash}X}
\toprule
Benchmark & Evaluated task set & Success criteria \\
\midrule
\revised{AdvancedIF~\citep{he2025advancedif}} & Prepared instruction-following tasks & All required instruction verdicts pass \\
\revised{CLBench DKR / RSA~\citep{dou2026cl}} & Domain Knowledge Reasoning / Rule System Application & Full-task rubric success \\
\revised{HealthBench~\citep{arora2025healthbench}} & Hard split, context-seeking and uncertainty themes & Normalized score at least $0.75$, plus negative-criterion and safety gates \\
\revised{ResearchRubrics \citep{sharma2025researchrubrics}} & Public processed research tasks & All mandatory criteria pass, with no penalties \\
\revised{MBPP Pro / HumanEval Pro~\citep{yu2025humanevalpro}} & $45$ / $40$ tasks after excluding defective tasks (Appendix~\ref{app:interface-audit}) & Executable test success \\
\bottomrule
\end{tabularx}
\caption{\textbf{Benchmarks and success criteria.} {The task set and success criterion of each benchmark.}}
\label{tab:detailed-benchmark-setup}
\end{table}

\subsection{Feedback, history, and grading}
\label{app:feedback-history-grading}

\paragraph{Feedback.}
Binary feedback tells the actor only that its attempt failed. Raw verifier feedback on AdvancedIF gives the verdict for each rubric item. Rubric-aware critics see the grading rubric when they write feedback, and rubric-blind critics do not. In coding, feedback either says that the code failed (binary) or shows the failed tests (test details), and Appendix~\ref{app:reference-protocol} gives the exact text. Raw verifier feedback and test details show the actor part of what is graded, so \seqk under these modes measures success with that information.

\paragraph{History.}
By default, the prompt for each sequential attempt contains the task, a note on the attempt number, and every earlier attempt with its feedback. In the summarized-history runs, a summary replaces the earlier attempts (Appendix~\ref{app:protocol-ablations}).

\paragraph{Temperature and repetitions.}
The main comparisons use $T=0.7$, and we do not tune the temperature. Coding runs are repeated three times, labeled seeds 1--3. Some coding runs in the history comparison use $T=0$.

\paragraph{Grading.}
Unless stated otherwise, AdvancedIF attempts are graded by Sonnet 4.6. CLBench attempts are graded by GPT-5.4, with GPT-5.2 as a fallback when GPT-5.4 refuses to grade.

\paragraph{Confidence intervals.}
We compute $95\%$ intervals by resampling tasks with replacement. Each resampled task keeps the results of every actor and protocol on it, so both sides of a comparison are always computed on the same tasks. Coding scores are averaged over the three repetitions before resampling. The model-selection analysis resamples tasks within each calibration and evaluation group (Appendix~\ref{app:budgeted-selection}).

\subsection{Coding prompts and feedback}
\label{app:reference-protocol}

Both coding benchmarks use the prompt below. The fields in braces are replaced with the benchmark's problem statements. The prompt does not contain the reference solution or the tests. Under \passk, every attempt receives this prompt unchanged, and all $K$ attempts are generated. Under \seqk, the run stops at the first passing attempt.
{\scriptsize
\begin{verbatim}
You are solving a Python code generation task.

There are two related functions:
1. Base function: solve the simpler base problem.
2. Main function: solve the harder problem by using or building on the base function.

Write complete Python code that defines all required functions.
Do not import unnecessary packages.
Do not include explanations outside the code.

Base problem:
```python
{base_problem}
```

Main problem:
```python
{main_problem}
```

Return only the Python code.
\end{verbatim}
}

Under \seqk, the task is followed by a note on the attempt number and then by every earlier attempt with its feedback, separated by blank lines. On the first attempt, the note is:
\begin{quote}\small\ttfamily
This is attempt 1 of \{K\}. If this attempt does not pass, you will receive feedback and can revise it on the remaining attempts.
\end{quote}
On later attempts, it is:
\begin{quote}\small\ttfamily
This is attempt \{t\} of \{K\}. Review your previous attempt(s) and the feedback below, then provide an improved answer.
\end{quote}
Each earlier attempt $i$ is added in order as follows. The whole prompt is sent as one user message. A variant of the note leaves out the number of attempts and keeps the rest.
{\small
\begin{verbatim}
<PreviousAttempt {i}>
{actor_output_i}
</PreviousAttempt {i}>

<Feedback {i}>
{feedback_i}
</Feedback {i}>
\end{verbatim}
}

Binary feedback is: ``Your previous solution failed the hidden tests. Please revise the code.'' Test-detail feedback begins with ``Your previous solution failed.'' and ``Failed test details:''. It then shows up to two failed tests, each with the function call, the expected output, and the actual output, or the error message when these are missing. Each field is cut at $300$ characters. The feedback ends with ``Use this feedback to fix the general logic. Do not hard-code these cases.'' and ``Return only the revised Python code.'' If the tests report no individual failures, the feedback is the raw test output. Some $T=0$ runs use error-summary feedback instead, which lists up to three failed tests with the test name, the error type, and a short summary, without the expected output.

\subsection{Excluded coding tasks}
\label{app:interface-audit}
In $15$ of the $100$ coding tasks, the tests call a function that the prompt never names, so an attempt can only pass if it guesses the name or learns it from the feedback. These are $10$ HumanEval Pro tasks (dataset IDs $3,5,14,20,21,22,28,46,63,112$) and $5$ MBPP Pro tasks (dataset IDs $19,41,47,69,75$). Among the HumanEval Pro tasks, seven give a base-function signature that does not match the tests, and three leave out the main function's definition. We find these tasks by checking the function names called in the tests against the names in the prompt, so other kinds of defects can remain.

The coding comparisons and the model-selection analysis exclude these tasks, which leaves $45$ MBPP Pro and $40$ HumanEval Pro tasks. The history comparison uses all $50$ tasks of each benchmark (Appendix~\ref{app:protocol-ablations}).

\section{Model Selection Details}
\label{app:selection-details}

This appendix gives the details of the selection test in Section~\ref{subsec:selection} and of the gaps between actors in Table~\ref{tab:separation-main}.

\subsection{Selection test}
\label{app:budgeted-selection}

\paragraph{Runs.}
The candidates are Sonnet 4.6, DeepSeek V3.2, Gemini 3 Flash, and Qwen3 Coder Flash. Each actor runs three times under \passk and three times under \seqk with test-detail feedback, at $T=0.7$ and $K=5$, on the $45$ MBPP Pro and $40$ HumanEval Pro tasks. This gives $48$ runs and $2{,}040$ trajectories. \seqthree and \seqfour are read from the first three or four attempts of the $K=5$ runs, whose prompts say that five attempts are available.

\paragraph{Selection and loss.}
On each set of calibration tasks, we score every actor by its success averaged over the three repetitions and pick the highest, breaking ties at random. The picked actor is evaluated by its test-detail \seqfive on the evaluation tasks, averaged over the three repetitions. Let $C$ be the calibration tasks, $E$ the evaluation tasks, $\mathcal M$ the candidate actors, $D_5(m,E)$ the \seqfive of actor $m$ on $E$, and $\widehat m(C)$ the actor picked on $C$. The selection loss is
\[
L(C,E)=\max_{m\in\mathcal M}D_5(m,E)-D_5(\widehat m(C),E).
\]
The evaluation tasks are never used for picking. The loss is zero when the picked actor is the best on $E$, and it can be positive even under \seqfive calibration, because the best actor on $C$ need not be the best on $E$. The best score on $E$ is the same for every calibration metric, so a lower loss means a higher \seqfive for the picked actor. The Gain in Table~\ref{tab:budgeted-selection-main} is this difference.

\paragraph{Cost.}
The calibration cost is the sum of the actor API charges over the four actors, the calibration tasks, and the three repetitions. Coding has no judge or critic, so this is the full model cost. Under both protocols, a trajectory is charged up to its first success or its last allowed attempt. The cost does not include running the tests. We report the mean cost and do not cap the spending on any single calibration set.

\paragraph{Calibration and evaluation groups.}
We split MBPP Pro into groups of $22$ and $23$ tasks and HumanEval Pro into two groups of $20$. We calibrate on one group, evaluate on the whole other group, and then swap. In each direction, we draw $128{,}000$ random orders of the calibration tasks and use the first $n$ tasks of each order, so every calibration metric sees the same tasks. We chose this number by increasing it from $2{,}000$ until two independent sets of draws agreed on the gain of \seqthree over \passfive at $n=10$ within $0.1$ points. The two directions are averaged, weighted by the size of their evaluation group.

\paragraph{Confidence intervals.}
{We compute $95\%$ intervals by resampling tasks with replacement $2{,}000$ times. Each resample draws tasks within each group and keeps the results of every actor, protocol, and repetition on a task together. On each resample, we rerun the selection over $200$ random orders of the calibration tasks in each direction. Using $400$ orders instead changes the interval endpoints by at most $0.3$ points. At $n=20$, HumanEval Pro calibrates on all $20$ tasks of its group, so the order of the tasks does not matter and only the resampling adds variation.}

\paragraph{\seqthree and \seqfour.}
Table~\ref{tab:budgeted-selection-prefix34} gives the gain and the cost relative to \passfive at $n=10$. We chose \seqthree before running the analysis and added \seqfour after seeing the \seqthree results. All four gain intervals exclude zero. \seqthree and \seqfour cost $0.94$ and $1.06$ times as much as \passfive on MBPP Pro, and $0.79$ and $0.92$ times as much on HumanEval Pro. Three of the four cost-ratio intervals include one, and on MBPP Pro, \seqthree costs more than \passfive on $28\%$ of the drawn calibration sets.

\begin{table}[!ht]
\centering\small\revisioncolor
\setlength{\tabcolsep}{3.5pt}
\caption{\textbf{\seqthree and \seqfour against \passfive with $10$ calibration tasks.} Cost ratio is the mean calibration cost of \seqk calibration divided by that of \passfive calibration. Gain is how much higher the picked actor's \seqfive is with \seqk calibration than with \passfive calibration, in percentage points. Brackets are $95\%$ bootstrap intervals.}
\label{tab:budgeted-selection-prefix34}
\begin{tabular}{lcrr}
\toprule
Benchmark & $K$ & Cost ratio [95\% CI] & Gain (pp) [95\% CI] \\
\midrule
MBPP Pro & 3 & $0.94 [0.83, 1.08]$ & $+8.3 [+1.8, +19.9]$ \\
MBPP Pro & 4 & $1.06 [0.90, 1.26]$ & $+8.1 [+1.5, +19.7]$ \\
HumanEval Pro & 3 & $0.79 [0.68, 0.91]$ & $+26.0 [+6.8, +44.7]$ \\
HumanEval Pro & 4 & $0.92 [0.76, 1.12]$ & $+25.9 [+7.0, +44.1]$ \\
\bottomrule
\end{tabular}
\end{table}

\paragraph{Scope.}
The gain of \seqthree over \passfive stays positive under all twenty other ways of splitting the tasks into two groups, on both benchmarks. These tasks are also used elsewhere in the paper. \seqthree and \seqfour come from runs that were told five attempts were available, so runs told they have three or four attempts may behave differently. We compare against \passfive only, not against every way of spending the same budget on independent samples.

\subsection{Gaps between actors}
\label{app:separation-uncertainty}

Table~\ref{tab:separation-main} reports two gaps for each protocol: the difference between the best and worst actor, and the average absolute difference over all pairs of actors. Table~\ref{tab:separation-ci} adds $95\%$ intervals for the change in each gap from \passfive to \seqfive. Each bootstrap draw resamples tasks, keeping all actors and both protocols of a task together, and recomputes both gaps. We use $2{,}000$ draws for the best--worst gap and $10{,}000$ for the all-pairs gap. Coding scores are averaged over the three repetitions before resampling, and AdvancedIF has one run per actor. The best--worst gap depends on which two actors are at the ends, so it moves more when actors are close. The all-pairs gap is less sensitive to this.

\begin{table}[!ht]
\centering\small
\caption{\textbf{Change in the gap between actors from \passfive to \seqfive, with $95\%$ intervals.} Values are in points, as in Table~\ref{tab:separation-main}.}
\label{tab:separation-ci}
\begin{tabular}{lrr}
\toprule
Benchmark & Best--worst gap change & All-pairs gap change \\
\midrule
AdvancedIF & $-26.7\ [-43.3,-3.3]$ & $-11.7\ [-21.0,-1.6]$ \\
MBPP Pro & $+12.6\ [-7.4,+29.6]$ & $+8.6\ [-2.8,+18.5]$ \\
HumanEval Pro & $+40.8\ [+12.5,+52.5]$ & $+25.4\ [+10.0,+31.9]$ \\
\bottomrule
\end{tabular}
\end{table}

\section{Cost Frontiers and Score Correlations}
\label{app:gain-correlation}

\subsection{Cost of each protocol}
\label{app:cost-replay}

Figure~\ref{fig:cost-comparison} compares the cost and success of \passk and \seqk for $K=1$ to $5$ without running new attempts. For each actor and task, we take the recorded $K=5$ runs of both protocols, go through their attempts in the original order, and stop at the first passing attempt or after attempt $K$. The cost of a task is the API charge of the attempts up to that point. Under \passk the actor never sees feedback, and the verdict is used only to decide when to stop. Each point in the figure is the mean cost per task and the success rate of one actor at one $K$.

On AdvancedIF, the cost includes the actor and the judge. On DKR, it includes the actor only.

All actors in a panel are scored on the same tasks: $30$ AdvancedIF tasks for seven actors, and $28$ DKR tasks for six actors. On DKR, Opus 4.7 and GPT-5.2 get directive feedback from themselves as critic, and the other actors get socratic feedback from GPT-5.2. Table~\ref{tab:dkr-common-ordering} gives the DKR scores on these $28$ tasks. Other DKR tables use $30$ tasks per actor, so their scores differ slightly. A point is on a protocol's frontier if no other point of that protocol reaches the same or higher success at a lower cost. The cost axis is logarithmic.

\begin{table}[!ht]
\centering\small\revisioncolor
\caption{\textbf{DKR scores on the $28$ tasks of Figure~\ref{fig:cost-comparison}.} Six actors, $T=0.7$, seed 42. The sequential critic differs across actors (Appendix~\ref{app:cost-replay}). Table~\ref{tab:correlation-scores} uses $30$ tasks and averages DeepSeek V4 Pro over four seeds, so its scores differ.}
\label{tab:dkr-common-ordering}
\begin{tabular}{lrr}
\toprule
Actor & Pass@5 & Seq@5 \\
\midrule
Opus 4.7 & 0.071 & 0.500 \\
DeepSeek V4 Pro & 0.143 & 0.429 \\
Gemini 3 Flash & 0.036 & 0.250 \\
GPT-5.2 & 0.143 & 0.286 \\
Qwen3.6 27B & 0.000 & 0.357 \\
Qwen3.6 Max & 0.071 & 0.393 \\
\bottomrule
\end{tabular}
\end{table}

\subsection{Correlation between protocols}

Figure~\ref{fig:gain-comparison} uses $26$ model--benchmark pairs: AdvancedIF (7), CLBench DKR (7), HealthBench (6), CLBench RSA (3), and ResearchRubrics (3). Coding is not included. Within a pair, both protocols use the same actor, temperature ($T=0.7$), $K=5$, grader, and tasks. When a pair has several sequential runs, we use the non-binary feedback setting with the full history that has the highest \seqfive. Because this choice is made on the same tasks, the \seqfive values are optimistic. Repeated runs of the same setting are pooled. Table~\ref{tab:correlation-scores} gives the scores of every pair.

Over the $26$ pairs, \passfive and \seqfive have Spearman correlation $+0.68$ and Pearson correlation $+0.81$. For the gains $\Delta\passfive$ and $\Delta\seqfive$, the Spearman correlation is $-0.08$ ($95\%$ interval $[-0.47,+0.31]$) and the Pearson correlation is $-0.04$ ($[-0.38,+0.26]$). The intervals come from $2{,}000$ bootstrap draws of the $26$ pairs, which treat the pairs as independent although actors and benchmarks recur. Eleven pairs have no gain under \passfive.

\begin{table}[!ht]
\centering\small\revisioncolor
\caption{\textbf{Scores of the $26$ model--benchmark pairs in Figure~\ref{fig:gain-comparison}.} $N$ is the number of shared tasks. All use $T=0.7$. \seqk uses the non-binary feedback setting with the highest \seqfive for each pair.}
\label{tab:correlation-scores}
\begin{tabular}{llrrrrr}
\toprule
Benchmark & Actor & $N$ & Pass@1 & Pass@5 & Seq@1 & Seq@5 \\
\midrule
AdvancedIF & Opus 4.7 & 30 & 0.433 & 0.633 & 0.467 & 1.000 \\
 & Sonnet 4.6 & 30 & 0.367 & 0.500 & 0.400 & 1.000 \\
 & DeepSeek V4 Pro & 30 & 0.433 & 0.667 & 0.567 & 0.967 \\
 & Gemini 3 Flash & 30 & 0.483 & 0.783 & 0.533 & 0.967 \\
 & GPT-5.2 & 30 & 0.400 & 0.550 & 0.400 & 0.933 \\
 & Qwen3.6 27B & 30 & 0.633 & 0.800 & 0.667 & 1.000 \\
 & Qwen3.6 Max & 30 & 0.600 & 0.833 & 0.667 & 1.000 \\
\midrule
DKR & Opus 4.7 & 30 & 0.000 & 0.067 & 0.000 & 0.500 \\
 & DeepSeek V4 Pro & 30 & 0.017 & 0.100 & 0.067 & 0.400 \\
 & Gemini 3 Flash & 30 & 0.000 & 0.033 & 0.033 & 0.267 \\
 & GPT-5.2 & 30 & 0.000 & 0.133 & 0.033 & 0.300 \\
 & GPT-5.5 Pro & 20 & 0.000 & 0.050 & 0.000 & 0.400 \\
 & Qwen3.6 27B & 29 & 0.000 & 0.000 & 0.000 & 0.345 \\
 & Qwen3.6 Max & 30 & 0.000 & 0.067 & 0.033 & 0.367 \\
\midrule
RSA & Opus 4.7 & 10 & 0.000 & 0.000 & 0.000 & 0.700 \\
 & Gemini 3 Flash & 10 & 0.000 & 0.000 & 0.000 & 0.300 \\
 & GPT-5.2 & 10 & 0.000 & 0.000 & 0.000 & 0.600 \\
\midrule
HealthBench & Opus 4.7 & 30 & 0.033 & 0.033 & 0.022 & 0.633 \\
 & DeepSeek V4 Pro & 30 & 0.067 & 0.067 & 0.011 & 0.632 \\
 & Gemini 3 Flash & 30 & 0.033 & 0.033 & 0.033 & 0.600 \\
 & GPT-5.2 & 30 & 0.050 & 0.075 & 0.033 & 0.800 \\
 & Qwen3 Max & 60 & 0.000 & 0.000 & 0.000 & 0.500 \\
 & Qwen3.6 Max & 30 & 0.000 & 0.033 & 0.011 & 0.611 \\
\midrule
ResearchRubrics & Opus 4.7 & 30 & 0.033 & 0.033 & 0.033 & 0.267 \\
 & Gemini 3 Flash & 30 & 0.000 & 0.000 & 0.000 & 0.067 \\
 & GPT-5.2 & 30 & 0.000 & 0.000 & 0.000 & 0.100 \\
\bottomrule
\end{tabular}
\end{table}

\section{Revision Dynamics Details}
\label{app:revision-details}

\subsection{Success by attempt}
\label{app:clbench-dkr-attempt-deltas}

Table~\ref{tab:per-attempt-deltas} gives \seqk for $K=1$ to $5$ on the DKR runs of Table~\ref{tab:clbench-dkr-feedback-gain}, with $30$ tasks per actor. Opus 4.7 and GPT-5.2 get directive feedback from themselves as critic, and DeepSeek V4 Pro, Gemini 3 Flash, and Qwen3.6 Max get socratic feedback from GPT-5.2. Since the feedback differs across actors, the rows do not compare actors under the same feedback.

\begin{table}[h]
\centering\small
\caption{\textbf{\seqk on DKR by attempt.} Seed 42, $30$ tasks per actor. The feedback setup differs across actors (Table~\ref{tab:clbench-dkr-feedback-gain}).}
\label{tab:per-attempt-deltas}
\begin{tabular}{lrrrrr}
\toprule
Actor & Seq@1 & Seq@2 & Seq@3 & Seq@4 & Seq@5 \\
\midrule
Opus 4.7 & 0.000 & 0.167 & 0.267 & 0.467 & 0.500 \\
DeepSeek V4 Pro & 0.067 & 0.167 & 0.333 & 0.367 & 0.400 \\
Gemini 3 Flash & 0.033 & 0.233 & 0.233 & 0.267 & 0.267 \\
GPT-5.2 & 0.033 & 0.100 & 0.133 & 0.300 & 0.300 \\
Qwen3.6 Max & 0.033 & 0.100 & 0.300 & 0.367 & 0.367 \\
\bottomrule
\end{tabular}
\end{table}

{When an actor revise successfully also depends on the feedback setup. In two other DKR runs with a different critic, grader, and note on the attempt number, Qwen3.6 Max has an EGS of $0.57$ and $0.50$, compared with $0.20$ in Table~\ref{tab:clbench-dkr-feedback-gain}. Final success changes as well. In runs where the actor critiques itself, GPT-5.4 grades, and the prompt leaves out the number of attempts, DeepSeek V4 Pro, Gemini 3 Flash, and Qwen3.6 27B reach \seqfive of $0.333$, $0.233$, and $0.267$. EGS and LGS therefore describe an actor under a given feedback setup, not the actor alone.}

\subsection{Annotation of revisions}
\label{app:repair-generalization}
{This section gives the details of the revision annotation in Section~\ref{subsubsec:revision-feedback}: which revisions we annotate, how the annotator labels them, and how often two annotators agree.}

\paragraph{Settings.}
We reuse recorded runs from four settings (Table~\ref{tab:repair-blocks}). Each setting fixes the feedback, critic, grader, temperature ($T=0.7$), and history, so actors within a setting differ only in the actor model and seed. A revision is one step from attempt $t$, through the feedback on it, to attempt $t{+}1$. A trajectory is included if its first attempt failed.

\paragraph{Task split.}
Tasks were split at random (seed $20260925$), so every run and attempt of a task falls on one side. The prompt was developed on the $14$ audited examples (Appendix~\ref{app:sec53-diagnostic-alignment}) and $11$ random tasks ($4$ AdvancedIF, $3$ DKR, $3$ HumanEval Pro, $1$ MBPP Pro). The two AdvancedIF settings share $16$ of their held-out tasks, giving $102$ distinct tasks: $60$ AdvancedIF, $23$ DKR, $11$ HumanEval Pro, and $8$ MBPP Pro. For coding we annotate $20$ random tasks out of $100$; four of them have interface defects (Appendix~\ref{app:interface-audit}). Figure~\panelref{fig:repair-trace}{b} excludes the $62$ revisions on those four tasks and the $131$ revisions whose feedback contains no valid request, leaving $842$.

\begin{table}[h]
\centering\footnotesize\setlength{\tabcolsep}{4pt}
\caption{Settings of the revision annotation. Tasks and revisions count held-out tasks only.}
\label{tab:repair-blocks}
\begin{tabular}{>{\raggedright\arraybackslash}p{0.17\linewidth}>{\raggedright\arraybackslash}p{0.25\linewidth}>{\raggedright\arraybackslash}p{0.32\linewidth}rr}
\toprule
Setting & Feedback; critic; grader & Actors (runs) & Tasks & Revisions \\
\midrule
AdvancedIF, o3-mini & raw verifier output; none; o3-mini & Gemini 3 Flash, GPT-5.2 ($K=5$); Opus 4.7, GPT-5.5 Pro ($K=3$) & $55$ & $153$ \\
AdvancedIF, Sonnet & raw verifier output; none; Sonnet 4.6 & $8$ actors, $13$ runs (Sonnet $\times5$, Haiku $\times2$) & $21$ & $256$ \\
CLBench DKR & rubric-aware socratic; GPT-5.2; GPT-5.2 & DeepSeek V4 Pro, Gemini 3 Flash, Qwen3.6 27B, Qwen3.6 Max & $23$ & $290$ \\
Coding & test details; none; tests & Sonnet 4.6, DeepSeek V3.2, Gemini 3 Flash, Qwen3 Coder Flash; $3$ seeds each & $19$ & $336$ \\
\bottomrule
\end{tabular}
\end{table}

\paragraph{Annotation.}
For each revision the annotator sees the task prompt, the rubric text, earlier attempts and feedback, attempt $t$, the feedback on it, attempt $t{+}1$, and a line diff. The actor, run, and all verdicts are hidden, and an annotation batch contains at most one revision per trajectory. The annotator lists each change the feedback requests and labels it valid, interpretive, unclear, or invalid (for example, a test that contradicts the prompt). It then records whether attempt $t{+}1$ satisfies each request (yes, partly, or no), with a verbatim quote as evidence. A request also counts as satisfied when attempt $t$ already met it and attempt $t{+}1$ still does. Whether a revision is unhelpful, addresses some, or addresses all of the feedback is computed from these labels after dropping invalid requests.

To develop the prompt, Claude Opus and Claude Sonnet independently labeled $68$ revisions from the development tasks ($16$ random revisions per setting and the two example AdvancedIF trajectories). The prompt was revised where they disagreed, then frozen before any held-out revision was annotated and before any verdict after attempt~1 was read.

\paragraph{Reliability.}
Claude Sonnet independently annotated a stratified $15\%$ of held-out revisions ($165$). The two annotators agree on this category for $80\%$ of them ($\kappa=0.73$); agreement is lowest in DKR ($\kappa=0.53$). A third blind Claude Opus pass resolved the $69$ disagreements; using its labels leaves the conclusions unchanged. All $2{,}029$ non-empty evidence quotes occur verbatim in the revised attempt. These are LLM labels, not human annotations, and three of the actors are Claude models.

\paragraph{Broken criteria.}
For the counts of revisions that address all feedback yet fail, a criterion counts as broken if it passed at attempt $t$, fails at attempt $t{+}1$, and no annotated request refers to it. Requests the annotator did not tie to a rubric item are treated as unmentioned, so these counts may slightly overstate breakage. Intervals in Figure~\panelref{fig:repair-trace}{b} are $95\%$ percentiles over $4{,}000$ resamples of tasks within each setting.

\section{Qualitative Analysis of Revision Trajectories}
\label{app:sec53-diagnostic-alignment}

\subsection{How we chose and read the examples}
We read $31$ trajectories ($115$ attempts) on $14$ tasks from AdvancedIF, CLBench DKR, HealthBench, and HumanEval Pro. We started from cases we noticed during the study, added runs of the same tasks under other feedback, and added one coding pair. Section~\ref{subsubsec:revision-feedback} shows two of these examples (Examples~\ref{card:sidehustle-gpt} and~\ref{card:spartan}), and this appendix shows two more. Table~\ref{tab:audited-trajectories} lists every trajectory we used. The examples show patterns of revision, not how often each pattern occurs. Appendix~\ref{app:repair-generalization} annotates revisions on $102$ other tasks.

For each trajectory we read the task, the attempts, the feedback, and the verdicts. A trajectory stops at its first passing attempt. $R_i$ is the rubric's own item number. Quotes are shortened to the relevant part. A contrast run starts from its own first attempt, so it does not show the same attempt under different feedback.

\subsection{{Revisions that do not act on the feedback: a tested function that is never defined}}
\label{app:qual-coding}
{This example shows an actor whose revisions do not act on clear feedback, including resubmitting the same code (Figure~\panelref{fig:repair-trace}{a}). Both actors get the same task and the same test-detail feedback, and only one of them fixes its code. The task is one of the defective tasks of Appendix~\ref{app:interface-audit}: the tests call a function that the prompt never names, so the feedback is the only way to learn the name. We chose this pair by taking the first task, by task ID, among the $17$ seed-1 tasks where both actors failed the first attempt, Sonnet 4.6 later passed, and Qwen3 Coder Flash never did.}
\begin{revexample}[label=card:coding, revcaption={Qwen3 Coder Flash under test-detail feedback, with Sonnet 4.6 on the same task and feedback as contrast.}]{HumanEval Pro}{Qwen3 Coder Flash; Sonnet 4.6}
\revfield{Task}{The task asks for a function that detects a negative running account balance and a main function that extends the check to several accounts. The prompt mistakenly repeats the base signature \texttt{below\_zero(operations: List[int])} for the main function, while the tests call \texttt{any\_account\_below\_zero}, which the prompt never names.}
\revfield{Feedback}{Deterministic test details, identical for both actors: ``Failed test details: - test\_1 function call: \texttt{any\_account\_below\_zero([[1, 2, 3], [4, 5, 6]])} \ldots\ Use this feedback to fix the general logic.''}
\revfield{Revision}{Qwen3 Coder Flash names its multi-account function \texttt{below\_zero}. Attempts 2 and 3 repeat attempt 1 unchanged. Attempt 4 notes the naming mismatch, calls it ``a copy-paste error in the main problem header'', and still keeps the old name; a header quoted without a body earlier in the same response also causes an \texttt{IndentationError}. Attempt 5 is identical to attempt 1.}
\revfield{Why Unhelpful}{The feedback shows the tested function name, but no revision defines it. Recognizing the mismatch in the explanation does not change the submitted code.}
\revfield{Contrast}{Sonnet 4.6 first names its multi-account function \texttt{below\_zero\_main}. Attempt 2 renames it to \texttt{any\_account\_below\_zero}, keeps its logic, and passes.}
\end{revexample}

\subsection{Revisions that break requirements that passed before: podcast review}
\label{app:qual-podcast}
This example shows how a revision that addresses the current feedback can break a requirement that passed before (Figure~\panelref{fig:repair-trace}{b}).
\revised{These podcast runs use socratic feedback with different critics and graders. They are separate from the directive runs for Opus~4.7 and GPT-5.2 in Table~\ref{tab:clbench-dkr-feedback-gain}.}
\begin{revexample}[label=card:podcast, revcaption={Podcast compliance: GPT-5.2, with the other actors' failed-requirement sets as contrast.}]{CLBench DKR}{GPT-5.2}
\revfield{Task}{The assistant reviews a podcast package for ethics and content safety under $18$ requirements. Among them: ensure statistics are current and sourced ($R_2$), flag damaging generalizations about groups ($R_4$), flag discriminatory ($R_8$) and dangerous ($R_9$) interview questions, refrain from fixing content with serious ethical concerns ($R_{11}$), prevent exposure of private information ($R_{13}$), and use the three-tier status system No Issue, Needs Revision, High Risk ($R_{16}$).}
\revfield{Feedback}{GPT-5.2 critiques its own review. After attempt 1, it asks whether the status labeling ``stay[s] consistent with the expected three-level scheme throughout the whole write-up'' ($R_{16}$) and whether any language ``paints a whole category of people'' ($R_4$). Later rounds also ask which elements are too high-stakes for ``script-ready replacement lines'' ($R_{11}$).}
\revfield{Revision}{The revisions adopt the three-tier labels, and $R_{16}$ passes by attempt 4. Attempt 5 fixes $R_4$ but proposes rewrites where the grader expects escalation, so $R_{11}$ fails.}
\revfield{Why Unhelpful}{Each revision fixes some flagged requirements while others fail. The failed sets are $\{4,8,11,16\}\to\{8,11,16\}\to\{4,16\}\to\{4\}\to\{11\}$, and the run never passes.}
\revfield{Contrast}{\begin{tabular}{@{}ll@{}}
Opus 4.7 & $\{11,16\}\to\{8\}\to\{9\}\to\emptyset$ \\
Qwen3.6 Max & $\{16\}\to\{2,11,13\}\to\emptyset$ \\
Qwen3.6 27B & $\{13\}\to\{16\}\to\emptyset$ \\
DeepSeek V4 Pro & $\emptyset$ at attempt 1 \\
\end{tabular}}
\end{revexample}
GPT-5.2 critiques itself and is graded by GPT-5.4. The number of failed requirements falls from $4$ to $1$, but which requirements fail keeps changing. $R_{16}$ passes from attempt 4, although the grader is more lenient about the labels than in earlier attempts. Opus 4.7 also critiques itself and is graded by GPT-5.4. Its second attempt adds a severity legend and labels each issue, which fixes $R_{16}$, and the remaining failure moves to $R_8$ and then $R_9$ before it passes. Qwen3.6 Max and Qwen3.6 27B use GPT-5.2 as critic and grader, and both pass at attempt 3. DeepSeek V4 Pro passes on its first attempt.

\subsection{List of trajectories}
Table~\ref{tab:audited-trajectories} lists every trajectory we used, including first-attempt successes and runs used only as contrasts. Unless noted, the non-coding runs use seed 42, $T=0.7$, and $K=5$. Opus 4.7's SideHustleBot run uses $K=3$, and Gemini 3 Flash's self-judge run on tinnitus uses $T=0$. The coding pair uses seed 1, $T=0.7$, full history, and test-detail feedback. In AdvancedIF, blind critique means that the actor critiques itself without the rubric. In the DKR podcast runs, Opus 4.7 and GPT-5.2 critique themselves, and DeepSeek V4 Pro and the Qwen models get feedback from GPT-5.2. The dog-summary and SpartanAI runs also use GPT-5.2 as critic. In HealthBench, self-judge feedback comes from the actor and judge feedback comes from the grader. Raw and binary feedback use no critic.
\begin{table}[H]
\centering\footnotesize
\begin{tabular}{lllll}
\toprule
Case / task index & Actor & Feedback & Judge & Attempts \\
\midrule
Hestia / 4 & Qwen Max & blind critique & GPT-5.4 & 2 \\
Hestia / 4 & Qwen Max & binary & GPT-5.4 & 5 \\
DJ / 12 & Gemini & blind critique & o3-mini & 5 \\
DJ / 12 & Gemini & raw & o3-mini & 2 \\
SideHustleBot / 13 & Qwen Max & raw & GPT-5.4 & 3 \\
SideHustleBot / 13 & Gemini & raw & o3-mini & 4 \\
SideHustleBot / 13 & Opus & raw & o3-mini & 1 \\
SideHustleBot / 13 & GPT-5.2 & raw & o3-mini & 5 \\
Channel notes / 7 & GPT-5.2 & aware socratic & GPT-5.4 & 5 \\
SpartanAI / 14 & DeepSeek & aware socratic & GPT-5.2 & 5 \\
Dog summary / 17 & DeepSeek & aware socratic & GPT-5.2 & 2 \\
Dog summary / 17 & DeepSeek & blind socratic & GPT-5.2 & 5 \\
Podcast / 20 & Opus & aware socratic & GPT-5.4 & 4 \\
Podcast / 20 & DeepSeek & aware socratic & GPT-5.2 & 1 \\
Podcast / 20 & GPT-5.2 & aware socratic & GPT-5.4 & 5 \\
Podcast / 20 & Qwen Max & aware socratic & GPT-5.2 & 3 \\
Podcast / 20 & Qwen 27B & aware socratic & GPT-5.2 & 3 \\
Knee / 1 & Gemini & judge feedback & GPT-5.4 & 2 \\
Cold feet / 16 & Opus & self-judge & GPT-5.4 & 2 \\
Cold feet / 16 & Opus & judge feedback & GPT-5.4 & 3 \\
Tinnitus / 22 & Gemini & raw & GPT-5.4 & 5 \\
Tinnitus / 22 & Gemini & self-judge & GPT-5.4 & 5 \\
Tinnitus / 22 & Gemini & judge feedback & GPT-5.4 & 5 \\
Mole / 25 & Qwen Max & self-judge & GPT-5.4 & 2 \\
Mole / 25 & Qwen Max & binary & GPT-5.4 & 5 \\
Fainting / 26 & Opus & self-judge & GPT-5.4 & 5 \\
Fainting / 26 & Opus & judge feedback & GPT-5.4 & 4 \\
Dosing / 28 & Opus & self-judge & GPT-5.4 & 5 \\
Dosing / 28 & Opus & judge feedback & GPT-5.4 & 5 \\
Coding interface / 4 & Sonnet & test details & Tests & 2 \\
Coding interface / 4 & Qwen Coder & test details & Tests & 5 \\
\bottomrule
\end{tabular}
\caption{\textbf{Trajectories we used.} The judge is the one set for the run. Task indices are positions within the run's benchmark, not dataset IDs.}
\label{tab:audited-trajectories}
\end{table}

\section{Feedback Loop Details}
\label{app:feedback-details}

\subsection{Criterion repair in DKR}
\label{app:sec53-binary}

Table~\ref{tab:per-model-feedback-flip-rate} breaks down the criterion repair of Table~\ref{tab:feedback-pooled} by actor. Each actor uses the $30$ tasks it has under all four conditions. For each task, we choose the attempt that passes the most criteria, taking the latest one if tied. Criterion repair is the share of criteria that fail in the first attempt but pass in the chosen attempt. Regression is the share of criteria that pass in the first attempt but fail in the chosen attempt. Pooled rates add up the criterion counts over actors before dividing. Each condition has its own first attempt, so the number of failed criteria differs across conditions. Regressions in attempts other than the chosen one are not counted. Binary feedback repairs $2.1$ points fewer criteria than independent sampling, with a $95\%$ interval of $[-11.5,+5.8]$ from $10{,}000$ resamples of the tasks (Appendix~\ref{app:feedback-history-grading}).

\begin{table}[!ht]
\centering
\caption{\textbf{Criterion repair and regression on DKR by actor (\%).} Each cell is repair / regression. Seed 42, $30$ tasks per actor. The rubric-aware critics are those of Table~\ref{tab:clbench-dkr-feedback-gain}, and the rubric-blind critic gives socratic feedback.}
\label{tab:per-model-feedback-flip-rate}
\small
\begin{tabular}{lrrrr}
\toprule
Actor & Independent & Binary & Rubric-blind critic & Rubric-aware critic \\
\midrule
Opus 4.7 & 37.2 / 1.0 & 36.9 / 0.7 & 26.0 / 1.9 & 77.1 / 2.3 \\
DeepSeek V4 Pro & 39.1 / 2.4 & 33.9 / 1.4 & 44.0 / 3.0 & 74.6 / 2.1 \\
Gemini 3 Flash & 30.8 / 4.2 & 28.5 / 3.1 & 42.0 / 3.5 & 69.8 / 3.0 \\
GPT-5.2 & 39.4 / 1.7 & 39.7 / 1.7 & 37.4 / 3.4 & 76.1 / 1.7 \\
Qwen3.6 Max & 37.9 / 2.3 & 34.8 / 2.4 & 30.7 / 2.6 & 72.2 / 1.3 \\
\midrule
Pooled & 36.5 / 2.2 & 34.4 / 1.8 & 36.8 / 2.9 & 73.8 / 2.1 \\
\bottomrule
\end{tabular}
\end{table}

\subsection{Binary feedback in coding}
\label{app:coding-results}

Table~\ref{tab:coding-numeric} gives the coding scores used in Table~\ref{tab:separation-main} and Section~\ref{subsec:informativeness}. Each cell averages three runs at $T=0.7$ on the $45$ MBPP Pro or $40$ HumanEval Pro tasks.

\begin{table}[h]
\centering\small
\setlength{\tabcolsep}{4pt}
\caption{\textbf{Coding scores with $K=5$.} Each cell averages three runs at $T=0.7$ on $45$ MBPP Pro or $40$ HumanEval Pro tasks.}
\label{tab:coding-numeric}
\begin{tabular}{llrrr}
\toprule
Benchmark & Actor & Pass@5 & Binary Seq@5 & Test-detail Seq@5 \\
\midrule
MBPP Pro & Sonnet 4.6 & 0.637 & 0.807 & 0.985 \\
MBPP Pro & DeepSeek V3.2 & 0.593 & 0.585 & 0.770 \\
MBPP Pro & Gemini 3 Flash & 0.526 & 0.622 & 0.978 \\
MBPP Pro & Qwen3 Coder Flash & 0.452 & 0.481 & 0.674 \\
\midrule
HumanEval Pro & Sonnet 4.6 & 0.358 & 0.617 & 0.975 \\
HumanEval Pro & DeepSeek V3.2 & 0.342 & 0.375 & 0.575 \\
HumanEval Pro & Gemini 3 Flash & 0.250 & 0.333 & 0.892 \\
HumanEval Pro & Qwen3 Coder Flash & 0.325 & 0.325 & 0.458 \\
\bottomrule
\end{tabular}
\end{table}

\Needspace{18\baselineskip}
\subsection{Summarized history}
\label{app:protocol-ablations}

Each unit is a pair of trajectories on the same task, one with a summarized history and one with the full history. The two trajectories come from two runs with the same benchmark, actor, feedback, budget, grader, prompt variant, seed, and temperature, and at least $30$ tasks in common. There are $57$ such run pairs, which give $2{,}520$ trajectory pairs. Table~\ref{tab:summary-history} reports the average change in success rate over the trajectory pairs of each benchmark. Unlike the other coding analyses, the coding rows use all $50$ tasks of each benchmark, including the defective tasks of Appendix~\ref{app:interface-audit}.

\begin{table}[H]
\centering\small
\caption{Change in success rate when the history is summarized, in percentage points, averaged over trajectory pairs. Each trajectory is scored at its run's attempt budget $K$. Coding runs use $T=0$, and the others use $T=0.7$.}
\label{tab:summary-history}
\begin{tabular}{lrrrrr}
\toprule
Benchmark & Trajectory pairs & Actors & Tasks & $K$ & Change (pp) \\
\midrule
HealthBench & 360 & 2 & 60 & 5 & $-14.4$ \\
CLBench DKR & 180 & 2 & 30 & 5 & $-6.1$ \\
ResearchRubrics & 180 & 4 & 30 & 5, 10 & $-7.2$ \\
AdvancedIF & 200 & 4 & 50 & 10 & $-0.5$ \\
MBPP Pro & 800 & 4 & 50 & 5, 10 & $+1.8$ \\
HumanEval Pro & 800 & 4 & 50 & 5, 10 & $+0.5$ \\
\bottomrule
\end{tabular}
\end{table}

\end{document}